\documentclass[10pt,twocolumn,twoside]{IEEEtran}
\usepackage[utf8]{inputenc}         
\usepackage[english]{babel}
\usepackage{multirow}

\usepackage{multicol}

\usepackage{soul}

\usepackage[normalem]{ulem}

\usepackage{xcolor,colortbl}

\ifCLASSINFOpdf
\usepackage[pdftex]{graphicx}
\usepackage{epstopdf}
\graphicspath{{./image/}}
\else
\fi
\usepackage[cmex10]{amsmath}
\usepackage{amssymb}
\usepackage{array}
\usepackage{rotating}

\DeclareMathOperator*{\softmax}{softmax}
\DeclareMathOperator*{\attention}{Attention}
\DeclareMathOperator*{\concat}{Concat}

\DeclareMathOperator*{\relu}{ReLU}
\DeclareMathOperator*{\multiheadattention}{MultiHeadAttention}
\DeclareMathOperator*{\layernorm}{LayerNorm}
\DeclareMathOperator*{\epochtransformer}{EpochTransformer}
\DeclareMathOperator*{\sequencetransformer}{SequenceTransformer}

\usepackage{makecell}
\usepackage{subcaption}
\usepackage{textcomp}

\definecolor{LightBlue}{rgb}{0.6172,0.7891,0.8789}

\usepackage{lipsum}

\usepackage{mdframed}

\usepackage{url}

\usepackage{cite}
\usepackage{capt-of}

\usepackage{tabularx}
\usepackage{bm}
\usepackage{calc}

\usepackage{xcolor,colortbl}
\usepackage{tabu}
\usepackage{mdframed}

\usepackage{pifont}

\usepackage{balance}

\begin{document}
	\title{SleepTransformer: Automatic Sleep Staging with Interpretability and Uncertainty Quantification}
	%
	%
	%

	\author{{Huy~Phan$^{*}$,
			Kaare Mikkelsen,
			Oliver~Y.~Ch\'{e}n,
			Philipp Koch,
			Alfred Mertins,
			and~Maarten~De~Vos
			\thanks{H. Phan is with the School of Electronic Engineering and Computer Science, Queen Mary University of London, London E1 4NS, UK and the Alan Turing Institute, London NW1 2DB, UK. K. Mikkelsen is with the Department of Electrical and Computer Engineering, Aarhus University, Aarhus 8200, Denmark. O. Y. Ch\'{e}n is with the School of Economics, Finance and Management, University of Bristol, Bristol BS8 1TU, UK. P. Koch and A. Mertins are with the Institute for Signal Processing, University of L\"ubeck, L\"ubeck 23562, Germany and with the German Research Center for Artificial Intelligence (DFKI), L\"ubeck 23562, Germany. M. De Vos is with the Department of Electrical Engineering and with the Department of Development and Regeneration, KU Leuven, 3001 Leuven, Belgium.}
			\thanks{$^*$Corresponding author: {\tt\footnotesize h.phan@qmul.ac.uk}}
	}}
	
	%
	%
	
	\markboth{THIS ARTICLE HAS BEEN PUBLISHED IN IEEE TRANSACTIONS ON BIOMEDICAL ENGINEERING}%
	{THIS ARTICLE HAS BEEN PUBLISHED IN IEEE TRANSACTIONS ON BIOMEDICAL ENGINEERING}
	%

	\IEEEpubid{DOI: 10.1109/TBME.2022.3147187 \hfill}


	\maketitle
	
	\begin{abstract}

		\emph{Background:} Black-box skepticism is one of the main hindrances impeding deep-learning-based automatic sleep scoring from being used in clinical environments. 
		\emph{Methods:} Towards interpretability, this work proposes a sequence-to-sequence sleep-staging model, namely SleepTransformer. It is based on the transformer backbone and offers interpretability of the model's decisions at both the epoch and sequence level. We further propose a simple yet efficient method to quantify uncertainty in the model's decisions. The method, which is based on entropy, can serve as a metric for deferring low-confidence epochs to a human expert for further inspection.
		\emph{Results:} Making sense of the transformer's self-attention scores for interpretability, at the epoch level, the attention scores are encoded as a heat map to highlight sleep-relevant features captured from the input EEG signal. At the sequence level, the attention scores are visualized as the influence of different neighboring epochs in an input sequence (i.e. the context) to recognition of a target epoch, mimicking the way manual scoring is done by human experts. 
		\emph{Conclusion:} Additionally, we demonstrate that SleepTransformer performs on par with existing methods on two databases of different sizes.
		\emph{Significance:} Equipped with interpretability and the ability of uncertainty quantification, SleepTransformer holds promise for being integrated into clinical settings.

	\end{abstract}
	

	%
	\IEEEpeerreviewmaketitle

	\section{Introduction}
	\label{sec:introduction}
	
	Sleep deprivation is prevalent and sleep disorders affect millions of people worldwide \cite{InstMed2006}, posing an huge burden on public health. Current practice of sleep diagnosis and assessment is still heavily dependent on human expertise. Machine intelligence, which is disrupting various application fields, holds huge potential for automating current sleep annotation. Although it is not the goal that machine intelligence will entirely replace human sleep experts \cite{Asan2020,Nagendran2020,Eysenbach2020}, we envision it could work alongside and assist human experts to facilitate their jobs and scale up sleep  assessment and diagnosis. 
	
	Sleep staging, the first and fundamental step in sleep diagnosis and assessment, is a typical application where machine intelligence can excel. In practice, this task of assigning a sleep stage to a 30-second sleep epoch is still being done manually, following a predefined set of rules, such as the American Academy of Sleep Medicine (AASM) guideline \cite{Iber2007}. On average, a sleep expert needs to spend two hours to complete scoring an overnight polysomnography (PSG) recording \cite{Malhotra2013}, making manually handling millions of sleep recordings infeasible. Automating this labor-intensive and routine process will free up a huge amount of time and efforts from sleep experts as a machine can complete the same task in a few seconds. Furthermore, automatic sleep scoring is indispensable when it comes to longitudinal sleep monitoring in home environments with novel mobile-EEG devices \cite{Mikkelsen2019, Mikkelsen2021}.
	
	\IEEEpubidadjcol
	Significant progress has been made towards automatic sleep staging in the last few years. The availability of large-scale public sleep databases with hundreds \cite{Oreilly2014} to thousands of subjects \cite{Zhang2018, Quan1997} has stimulated and enabled the exploration of deep learning paradigms in solving this problem \cite{Stephansen2018, Supratak2017, Biswal2018a, Phan2019a, Phan2021c, Chambon2018}. Early attempts tried to use vanilla deep network architectures, such as deep neural networks (DNNs) \cite{Dong2017}, convolutional neural networks (CNNs) \cite{Chambon2018, Tsinalis2016, Sun2017, phan2018c, Sors2018}, and recurrent neural networks (RNNs) \cite{phan2018d}. Replacing more conventional machine learning methods with these vanilla networks in simple one-to-one or many-to-one frameworks resulted in limited success, owing to the limitation of the short input context. Since the seminal work in \cite{Phan2019a}, the sequence-to-sequence sleep staging approach has grown in popularity for the task. Using this framework,  handling a long context of 20-30 consecutive PSG epochs simultaneously, various advanced architectures have been proposed, for example CNN+RNN \cite{Seo2020, MousaviI2019}, hierarchical RNN \cite{Phan2019a,Guillot2021}, and CNN+Transformer \cite{Fan2021}. It further allows  extensions from different angles, such as transfer learning \cite{Phan2020b,Phan2019c, Banluesombatkul2021}, model personalization \cite{Phan2020a, Mikkelsen2018}, and multi-view learning \cite{Phan2021c}. These advances have significantly pushed the performance of machine sleep scoring to be on par with human scoring \cite{Stephansen2018,Phan2019a,Guillot2021,Phan2021c}.
	
	Despite all this progress, we have not yet seen automatic sleep staging widely adopted clinically. Unofficial communications with leading sleep experts point to the scepticism of deep learning models being a black box, which is a common criticism when it comes to the application of artificial intelligence in healthcare and medicine \cite{Amann2020}. We argue that two overarching obstacles need to be addressed for a machine scoring system to work alongside practitioners in an interactive and collaborative manner: (1) interpretability \cite{Vilamala2017,Lee2020} and (2) uncertainty quantification \cite{Mikkelsen2020}. 
	Interpretability is the ability of a model to explain how its decision is made given a certain input, to be understood by a human. Inspired by the way a sleep expert performs manual scoring \cite{Iber2007}, interpretability in automatic sleep scoring is reasonably about (but not limited to) what features the model learns from the input signal, whether these features are relevant to and underpin the sleep stages, and how the decision on a target epoch is made under the influence of its neighboring epochs. Interpretability is particularly important due to the fact that sleep stages are ambiguous and even different human experts tend to disagree at a certain extend \cite{Guillot2020,DankerHopfe2009}. Also, due to this ambiguity, quantifying uncertainty in the model's decisions is equally important. 
	Simply put, we are in need of a simple and concrete metric, ideally a single number, for quantifying the model's uncertainty. Using this metric, epochs that are scored with low confidence by the model can be deferred to sleep experts for further inspection \cite{Becker2021}. 
	
	In this work, we propose a sleep staging model, namely SleepTransformer, as a stepping stone towards addressing the two above-mentioned obstacles. SleepTransformer adheres to the sequence-to-sequence sleep staging framework \cite{Phan2019a,Phan2020b}. However, different from most (if not indeed all) existing works, SleepTransformer is convolution- and recurrent-free. Instead, it relies on the transformer concept \cite{Vaswani2017} as the backbone for both epoch- and sequence-level modelling. The transformer construction is solely based on a self-attention mechanism whose attention scores will be leveraged for the model's interpretability at both the epoch and sequence level. On the one hand, the attention scores at the epoch level will be used as a heat map applied to the EEG signal input to highlight the features the model attends to. On the other hand, the attention scores at the sequence level is interpreted as the influence of different neighboring epochs to the recognition of a target epoch in an input sequence. We also propose to use entropy of the multi-class probability distribution outputted by the model to neatly quantify uncertainty in its decisions. We show that the estimated uncertainty allows us to identify most of the model's mistakes. Experimental results on two public databases, Sleep Heart Health Study (SHHS) and Sleep-EDF Expanded, of varying size also show that SleepTransformer performs comparably to existing state-of-the-art models on the two databases.
	
	Our major contributions are summarized as follows.
	
	\begin{itemize}
		\item The proposed SleepTransformer is a transformer-based sequence-to-sequence model which achieves state-of-the-art performance on automatic sleep scoring. To the best of our knowledge, this is the first sequence-to-sequence model solely relying on the transformer architecture proposed for the task.
		\item We address interpretability of a sleep-staging model in a natural way at both the epoch and sequence level by leveraging the attention scores of the transformer's self-attention module.
		\item We propose an entropy-based method to elegantly quantify uncertainty in the model's decisions as a concrete number.
	\end{itemize}
	
	

	
	The rest of the article is organized as follows. We outline the used databases in Section \ref{sec:materials}. We then describe the architecture of the transformer backbone in Section \ref{ssec:transformer}, followed by the proposed SleepTransformer in Section \ref{sec:sleeptransformer}. We elaborate the interpretability and uncertainty quantification of the model in Section \ref{sec:explainability_and_confidence}. Details about the
	experiments will be presented in Section \ref{sec:experiments}. We conclude the article in Section \ref{sec:conclusions}.
	
	\section{Materials}
	\label{sec:materials}

	The following two databases will be used for experiments in this work:
	
	{\bf  Sleep Heart Health Study (SHHS):} This is a large-scale database collected from multiple centers to study the effect of sleep-disordered breathing on cardiovascular diseases \cite{Zhang2018, Quan1997}. The data was collected as part of the clinical trial ``Sleep Heart Health Study (SHHS)'', ClinicalTrials.gov number, NCT00005275. It has two rounds of PSG records, namely Visit 1 (SHHS-1) and Visit 2 (SHHS-2). The former, consisting of 5,791 subjects aged 39-90, was employed in this work. Manual scoring was completed using the R\&K guideline \cite{Hobson1969}. Similar to other databases annotated with the R\&K rule, N3 and N4 stages were merged into N3 stage and MOVEMENT and UNKNOWN epochs were discarded. We adopted C4-A1 EEG in the experiments.
	
	{\bf SleepEDF-78:} This database is the 2018 version of the Sleep-EDF Expanded dataset \cite{Kemp2000, Goldberger2000}, consisting of 78 healthy Caucasian subjects aged 25-101. Two consecutive day-night PSG recordings were collected for each subject, except subjects 13, 36, and 52 whose one recording was lost due to device failure. Manual scoring was done by sleep experts according to the R\&K standard \cite{Hobson1969} and each 30-second PSG epoch was labeled as one of eight categories \{W, N1, N2, N3, N4, REM, MOVEMENT, UNKNOWN\}. N3 and N4 stages were merged into N3 stage. MOVEMENT and UNKNOWN epochs were excluded. We used the Fpz-Cz EEG in this study. Of note, we adhere to the common setting where a recording was trimmed starting from 30 minutes before to 30 minutes after its \emph{in-bed} part \cite{Phan2021c}.
	
	\section{Transformer}
	\label{ssec:transformer}
	Transformer \cite{Vaswani2017}, a sequence model solely based on self-attention, has shown compelling results on various sequential modelling tasks. The transformer is composed of an encoder and a decoder sharing the same model architecture. However, the decoder is a left-context-only	version which is tasked for generation purpose. To avoid confusion, the transformer used in this work is the encoder part.  It comprises two core modules: 
	multi-head attention and position-wise feed-forward network. 
	
	
	\begin{figure} [!b]
		\centering
		\includegraphics[width=1\linewidth]{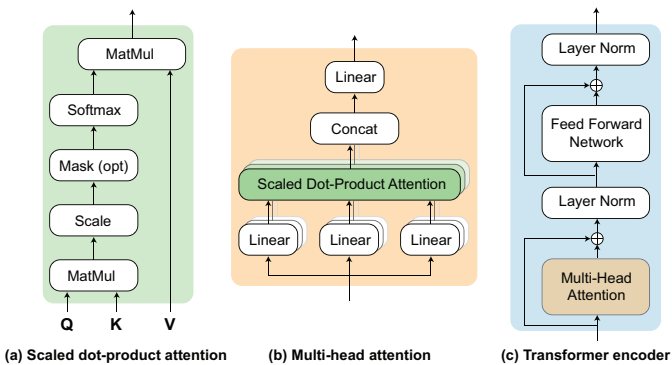}
		\caption{Architecture of (a) scaled dot-product attention, (b) multi-head attention, and (c) transformer encoder.}
		\label{fig:components}
	\end{figure}
	
	The attention mechanism used in the multi-head attention module is scaled dot-product attention, as illustrated in Figure~\ref{fig:components}~(a). It associates elements at different positions of an input sequence to derive the output sequence which is computed as a weighted sum of the input values, where the weight for each value is computed by an attention function of the query with the corresponding keys. Multi-head attention is composed of $H$ scaled dot-product attention modules, as illustrated in Figure~\ref{fig:components} (b). Firstly, $H$ different learnable linear projections are applied to the input and map it to parallel queries, keys, and values. Then, the scaled dot-product attention is performed on these mapped queries, keys, and values simultaneously. The $H$ attention heads are then concatenated, followed by a linear projection to produce the attentive output. All these steps can be formulated as follows:
	\begin{align}
		\mathbf{Q}_i = \mathbf{Z}\mathbf{W}_i^Q, \text{~}\mathbf{K}_i &= \mathbf{Z}\mathbf{W}_i^K, \text{~}\mathbf{V}_i = \mathbf{Z}\mathbf{W}_i^V, \text{~} 1 \le i \le H,
		\label{eq:matrices} \\
		\mathbf{H}_i &= \attention(\mathbf{Q}_i, \mathbf{K}_i, \mathbf{V}_i) \nonumber \\ 
		&= \softmax(\frac{\mathbf{Q}_i\mathbf{K}_i^{\mathsf{T}}}{\sqrt{d}})\mathbf{V}_i,
		\label{eq:attention} \\
		\tilde{\mathbf{Z}} &= \concat(\mathbf{H}_1, \ldots, \mathbf{H}_H)\mathbf{W}^Z.
		\label{eq:multihead_attention}
	\end{align}
	Here, $\mathbf{Z} \in \mathbb{R}^{l\times d}$ is the input with length $l$ and dimension $d$. $\mathbf{Q}_i, \mathbf{K}_i, \mathbf{V}_i \in \mathbb{R}^{l\times \frac{d}{H}}$ are the mapped queries, keys, and values. $\mathbf{H}_i \in \mathbb{R}^{l\times \frac{d}{H}}$ is the $i$-th attention head. $\mathbf{W}_i^Q, \mathbf{W}_i^K, \mathbf{W}_i^V \in \mathbb{R}^{d\times \frac{d}{H}}$ and $\mathbf{W}^Z \in \mathbb{R}^{d\times d}$ are the learnable weight matrices. $\tilde{\mathbf{Z}} \in \mathbb{R}^{l\times d}$ is the attentive output.
	
	The position-wise feed-forward network is a fully connected feed-forward network. It is comprised of two linear transformations with a ReLU activation in between. Besides the two main modules, the transformer also includes several residual and normalization layers as illustrated in Figure~\ref{fig:components} (c). As a whole, it can be formulated as follows:
	\begin{align}
		\tilde{\mathbf{Z}} &= \multiheadattention(\mathbf{Z}), \\
		\mathbf{Z}_{mid} &= \layernorm(\mathbf{Z} + \tilde{\mathbf{Z}}), \\
		\mathbf{Z}_{FF} &= \relu(\mathbf{Z}_{mid}\mathbf{W}_1 + \mathbf{b}_1)\mathbf{W}_2 + \mathbf{b}_2, \\
		\mathbf{O} &= \layernorm(\mathbf{Z}_{mid} + \mathbf{Z}_{FF}).
	\end{align}
	Here, $\mathbf{Z}_{FF}$ denotes the output of the position-wise feed-forward network, in which $\mathbf{W}_1\!\in\!\mathbb{R}^{d\times d_{FF}}$, $\mathbf{W}_2\!\in\!\mathbb{R}^{d_{FF}\times d}$, $\mathbf{b}_1\!\in\!\mathbb{R}^{d_{FF}}$, and $\mathbf{b}_2\!\in\!\mathbb{R}^{d}$ are learnable weight matrices and biases, respectively. 
	
	%
	%
	%
	
	\begin{figure} [!t]
		\centering
		\includegraphics[width=1\linewidth]{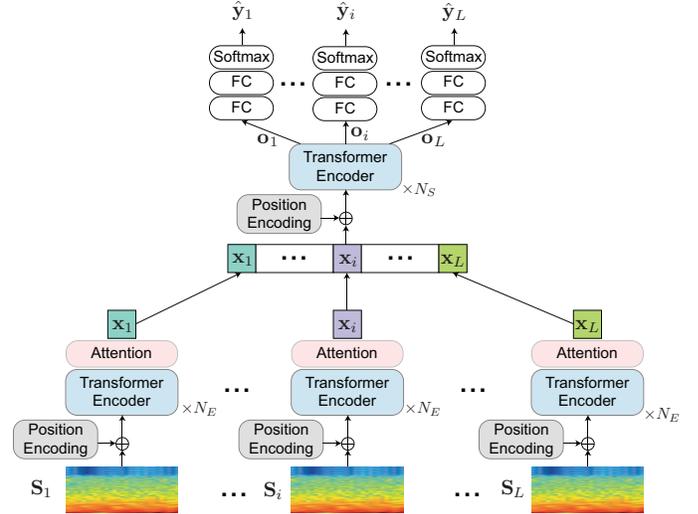}
		\caption{Illustration of SleepTransformer.}
		\label{fig:sleeptransformer}
		\vspace{-0.3cm}
	\end{figure}
	\section{SleepTransformer}
	\label{sec:sleeptransformer}
	
	Given a training set $\{\mathcal{S}_n\}_{n=1}^N$ of size $N$ where $\mathcal{S}_n = (\{\mathbf{S}^{(n)}_1, \mathbf{y}^{(n)}_1\}, \ldots, \{\mathbf{S}^{(n)}_L, \mathbf{y}^{(n)}_L\})$ is the $n$-th sequence of $L$ sleep epochs. $\mathbf{S}^{(n)}_i\!\in\!\mathbb{R}^{T\times F}$, $1\le\!i\!\le\!L$ represents a time-frequency image of $T\!=\!29$ time frames and $F\!=\!128$ frequency bins extracted from the $i$-th 30-second EEG epoch in the $n$-th sequence (see Section \ref{ssec:featureextraction}). $\mathbf{y}^{(n)}_i \in \{0,1\}^C$ denotes the one-hot encoding label of the $i$-th EEG epoch in the $n$-th sequence, where $C=5$ as we are dealing with 5-stage sleep staging.
	
	The proposed SleepTransformer, illustrated in Figure \ref{fig:sleeptransformer}, adheres to the sequence-to-sequence sleep staging framework proposed in \cite{Phan2019a}. It uses the transformer described in Section \ref{ssec:transformer} as the backbone network for both intra-epoch (i.e. epoch level) and inter-epoch (i.e. sequence level) processing. It is, therefore, free of convolutional and recurrent components which are the main ingredients in existing deep-learning models for sleep staging, such as \cite{Phan2019a,Supratak2017,Phan2021c,Chambon2018,Olesen2021,Stephansen2018,Perslev2019,Perslev2021}. 
	
	\subsection{Epoch transformer}
	\label{ssec:epochtransformer}
	The epoch transformer plays the role of a feature map that transforms a 30-second EEG epoch into a feature vector for representation. This feature map has been commonly realized either by a CNN \cite{Supratak2017,Phan2021c} or an RNN \cite{Phan2019a}. Orthogonally, SleepTransformer realizes this map using $N_E$ transformers.
	
	A time-frequency image $\mathbf{S}$ is treated as a sequence of $T$ spectral columns. Without confusion, we omit the superscript and subscript for simplicity. 
	We aim to encode this sequence\ by a heap of $N_E$  transformers which are denoted as \emph{EpochTransformer}. As the transformer itself cannot encode the order information which is vital for both intra-epoch and inter-epoch processing in a sequence-to-sequence sleep-scoring network \cite{Phan2019a}, we firstly add positional encodings to the input to introduce order information:
	\begin{align}
		\mathbf{\tilde{S}} = \mathbf{S} + \mathbf{P}^\text{ep}. \label{eq:SplusP}
	\end{align}
	In (\ref{eq:SplusP}), $\mathbf{P}^\text{ep} \in \mathbb{R}^{T\times F}$ denotes the positional encoding matrix. We use sine and cosine functions as in the seminal work \cite{Vaswani2017} for positional encoding purpose where the $i$-th row and the $(2j)$-th or the $(2j+1)$-th column is given as 
	\begin{align}
		p_{i,2j} = \sin\left(\frac{i}{10000^{2j/F}}\right), \label{p2j} \\
		p_{i,2j+1} = \cos\left(\frac{i}{10000^{2j/F}}\right). \label{p2jplus}
	\end{align}
	The sequence of spectral columns of $\mathbf{\tilde{S}}$ is then modelled by  
	\begin{align}
		\mathbf{X}^{(i)} = \epochtransformer\left(\mathbf{X}^{(i-1)}\right), \label{eq:epochtransformer}
	\end{align}
	where $\mathbf{X}^{(i)}\!\in\!\mathbb{R}^{T\times F}$,  $1\!\le\!i\!\le\!N_E$, and $\mathbf{X}^{(0)}\!\equiv\!\mathbf{\tilde{S}}$. In order to reduce $\mathbf{X}^{(N_E)}$, the output of the last transformer in the heap, to a compact feature vector for epoch-wise representation, we combine its columns $(\mathbf{x}^{(N_E)}_1, \ldots, \mathbf{x}^{(N_E)}_T)$ via a weighted combination:
	\begin{align}
		\mathbf{x} = \sum\nolimits_{t=1}^T \alpha_t\mathbf{x}^{(N_E)}_t. \label{eq:attention_feat}
	\end{align}
	In (\ref{eq:attention_feat}), $\mathbf{x}\in\mathbb{R}^{F}$ denotes the derived feature vector that represents the input epoch. $\alpha_1, \ldots, \alpha_T$ are the attention weights learned by a softmax attention layer as in \cite{phan2018d,Phan2019a}:
	\begin{align}
		\alpha_t &= \frac{\exp(\mathbf{a}_t^{\mathsf{T}}\mathbf{a}_e)}{\sum_{t=1}^{T}\exp(\mathbf{a}_t^{\mathsf{T}}\mathbf{a}_e)}, \label{eq:attention_weights} \\ \mathbf{a}_t &= \tanh(\mathbf{W}_a\mathbf{x}_t + \mathbf{b}_a),
	\end{align}
	where $\mathbf{W}_a\in \mathbb{R}^{A\times F}$ and $\mathbf{b}_a\in \mathbb{R}^{A}$ are a learnable weight matrix and bias, respectively. $\mathbf{a}_e\in \mathbb{R}^{A}$ is the trainable epoch-level context vector. $A$ is the so-called attention size.
	\subsection{Sequence transformer}
	\label{ssec:sequencetransformer}
	
	Via the epoch transformer in Section (\ref{ssec:epochtransformer}), an input sequence $(\mathbf{S}_1, \ldots, \mathbf{S}_L)$ has now been transformed into a sequence of epoch-wise feature vectors $(\mathbf{x}_1, \ldots, \mathbf{x}_L)$, where $\mathbf{x}_i, 1\!\le\!i\!\le\!L$, is given in (\ref{eq:attention_feat}). In existing work complying to the sequence-to-sequence sleep staging framework, the resulting epoch-wise feature vectors $(\mathbf{x}_1, \ldots, \mathbf{x}_L)$ were typically processed by a bidirectional RNN for inter-epoch modelling \cite{Phan2019a}. Here, we employ a heap of $N_S$ transformers, denoted as \emph{SequenceTransformer}, for this purpose.
	
	Similar to the epoch transformer, positional encoding is firstly carried out via sine-and-cosine functions:
	\begin{align}
		\mathbf{\tilde{X}} = \mathbf{X} + \mathbf{P}^\text{seq}, \label{eq:SplusP_sequence}
	\end{align}
	where $\mathbf{X} = (\mathbf{x}_1, \ldots, \mathbf{x}_L) \in \mathbb{R}^{L\times F}$. $\mathbf{P}^\text{seq} \in \mathbb{R}^{L\times F}$ denotes the positional encoding matrix whose elements are computed using (\ref{p2j}) and (\ref{p2jplus}).  $\mathbf{\tilde{X}}$ is then processed by the heap of $N_S$ \emph{SequenceTransformer}:
	\begin{align}
		\mathbf{O}^{(i)} = \sequencetransformer\left(\mathbf{O}^{(i-1)}\right), \label{eq:sequencetransformer}
	\end{align}
	where $\mathbf{O}^{(i)} \in \mathbb{R}^{L\times F}$,  $1 \le i \le N_S$, and $\mathbf{O}^{(0)} \equiv \mathbf{\tilde{X}}$.
	
	Given the output of the last \emph{SequenceTransformer}, $\mathbf{O}^{(N_S)} = (\mathbf{o}^{(N_S)}_1, \ldots, \mathbf{o}^{(N_S)}_L)$, the vectors $\mathbf{o}^{(N_S)}_i$, $1 \le i \le L$, are eventually presented to two fully-connected (FC) layers with ReLU activation, followed by a softmax layer to obtain the output sequence $(\mathbf{\hat{y}}_1, \ldots, \mathbf{\hat{y}}_L)$. As in \cite{Phan2019a,Phan2021c}, SleepTransformer is trained to minimize the cross-entropy loss over the sequence: 
	\begin{align}
		\mathcal{L} = -\frac{1}{L}\sum\nolimits_{i=1}^L\mathbf{y}_i\log(\hat{\mathbf{y}}_i).
	\end{align}
	
	\section{Interpretability  and Confidence Quantification}
	\label{sec:explainability_and_confidence}
	\subsection{Interpretability via self-attention}
	\label{ssec:explainability}
	
	Self-attention (cf. Figure \ref{fig:components} (a)) learns a representation by relating the input elements at different positions in the input sequence. From the given query $\mathbf{Q}$, the machine learns the relation between the query and keys $\mathbf{K}$ to compute attention scores, and multiply the attention scores to the values $\mathbf{V}$. Finally, the sum of attended values composes the semantics of the given query. The attention scores can be leveraged to interpret the model. We propose two different visualizations for interpretation: (1) EEG attention heat map that shows where in the input EEG signal the model pays more attention to, and (2) epoch influence as a bar chart which qualifies the contribution of neighboring epochs to predicting the sleep stage of a target epoch in the input sequence.
	
	{\bf EEG attention heat map.} To understand the behavior of the model, it is important to know what parts of the EEG input the model pays more attention to. To this end, we sum the attention scores from each attention head of \emph{EpochTransformer} for visualization. Attention score $\mathbf{A}$ of a single attention head is given as:
	\begin{align}
		\mathbf{A} = \softmax\left(\frac{\mathbf{Q}\mathbf{K}^\mathsf{T}}{\sqrt{d}}\right),
	\end{align}
	where $\mathbf{A}\!\in\!\mathbb{R}^{l\times l}$. The element $a_{i,j}$ at $i$-th row and $j$-th column indicates how much the input at index $j$ attributes to the representation at index $i$. The matrix $\mathbf{A}$ is, therefore, summed in the first dimension, followed by normalization to the range $[0,1]$ to obtain a score vector whose $i$-th element indicates how much the input at index $i$ attributes to the representations at all other indices. 
	
	A potential pitfall of the above-mentioned heat map is that summing the attention scores across the attention heads may dismiss attention structures of the attention heads. As an alternative, to further gain insight into the representation learned by the network, we use the attention score matrices to transform a time-frequency input and obtain a time-frequency output right after the last \emph{EpochTransformer}, omitting all other non-linear operations. Inverse short-time Fourier Transform (ISTFT) is then applied the time-frequency output to construct the raw EEG signal which can be visualized to exhibit the features learned by the network. Of note, we use the original phase of the time-frequency input for this construction.
	
	{\bf Epoch influence bar chart.} To further shed light on the behavior of the model, it is equally important to know which neighboring epochs the model pays more attention to while scoring the target epoch in the input sequence. Given the attention score matrix $\mathbf{A}$ of a \emph{SequenceTransformer}, the element $a_{i,j}$ at $i$-th row and $j$-th column indicates how much the epoch $j$ in the input sequence is attributing to the representation of the target epoch $i$. We argue that it closely resembles the way a clinician performs manual scoring. Specifically, when the target epoch does not show much evidence of sleep-relevant features, the neighboring epochs in the context will be attended to, providing evidence in support of the scoring \cite{Iber2007}.

	\subsection{Entropy-based confidence quantification}
	\label{ssec:confidence_estimation}
	
	In a general multi-class classification problem, a deep neural network outputs a vector whose elements are probabilities, one for each target class of interest. For the 5-stage sleep staging we are dealing with, an output $\mathbf{\hat{y}}$ from SleepTransformer consists of $C$ probability values ($C=5$ in this case) corresponding to $C$ sleep stages. Typically, the sleep stage with respect to the maximum probability is considered the network's prediction. However, the predicted discrete label does not tells us how much the network is confident about its decision whereas the multi-class probability distribution $\mathbf{\hat{y}}$ is too complex.
	
	
	In fact, the multi-class probability distribution over the sleep stages encoded in $\mathbf{\hat{y}}$ can provide a more refined measure of confidence in the network prediction. In one extreme, when $\mathbf{\hat{y}}$ assigns probability 1 to one class and  probability 0 to the remaining classes, we expect the network to be very confident in its decision. In the other extreme, when the distribution is flat, i.e. all elements in $\mathbf{\hat{y}}$ are equal, the network has no confidence in its decision. All other distributions indicate varying levels of confidence between these two extremes. The entropy of the discrete probability distribution, an information-theoretic measure of uncertainty \cite{Cover2006}, appears to be a natural way to measure the network's intrinsic uncertainty. In turn, the network's confidence can be quantified as a concrete number. To this end, we propose to use normalized entropy: 
	\begin{align}
		H(\mathbf{\hat{y}}) = -\sum\nolimits_{c=1}^C \hat{y}_c\frac{\log(\hat{y}_c)}{\log C}, \label{eq:entropy}
	\end{align}
	to normalize the range of the uncertainty to $[0, 1]$, assuming $0\!\times\!\log0 = 0$. In turn, the network confidence is quantified as
	\begin{align}
		Conf(\mathbf{\hat{y}}) =  1 - H(\mathbf{\hat{y}}). \label{eq:confidence}
	\end{align}
	For 5-stage classification, $H(\mathbf{\hat{y}}) = 1$ and $Conf(\mathbf{\hat{y}}) = 0$ when $\mathbf{\hat{y}} = (\frac{1}{5}, \frac{1}{5}, \frac{1}{5}, \frac{1}{5}, \frac{1}{5})$. $H(\mathbf{\hat{y}}) = 0$ and $Conf(\mathbf{\hat{y}}) = 1$ when $\mathbf{\hat{y}}$ contains exactly one probability 1, for example $\mathbf{\hat{y}} = (0, 1, 0, 0, 0)$. All other possible values of $\mathbf{\hat{y}}$ will result in $0 < Conf(\mathbf{\hat{y}}) < 1$.

	Given the estimated confidence, we envision that a low-confidence epoch can be deferred for further manual verification and correction by human experts. The filtering can be accomplished via either thresholding the confidences with a predefined threshold or simply selecting a certain percentage of epochs with lowest confidences. 
	
	
		\setlength\tabcolsep{2.25pt}
	\begin{table*}[!t]
		\caption{Performance comparison between SleepTransformer and previous works on the experimental databases. The superscript $^*$ indicates the model was initialized by the model pretrained on the SHHS database. $^\dagger$ The results are not directly comparable either due to the differences in the data split and the channels used (Olesen \emph{et al.} \cite{Olesen2021} and U-Sleep \cite{Perslev2021}) or due to the use of a small subset of healthy subjects (Eldele \emph{et al.} \cite{Eldele2021}).}
		\vspace{-0.2cm}
		\begin{center}
			\begin{tabular}{|>{\centering\arraybackslash}m{0.75in}|>{\arraybackslash}m{1in}|>{\centering\arraybackslash}m{0.4in}|>{\centering\arraybackslash}m{0.45in}|>{\centering\arraybackslash}m{0.4in}|>{\centering\arraybackslash}m{0.4in}|>{\centering\arraybackslash}m{0.4in}||>{\centering\arraybackslash}m{0.4in}|>{\centering\arraybackslash}m{0.45in}|>{\centering\arraybackslash}m{0.4in}|>{\centering\arraybackslash}m{0.4in}|>{\centering\arraybackslash}m{0.4in}|>{\centering\arraybackslash}m{0in} @{}m{0pt}@{}}
				\cline{1-12}
				\multirow{2}{*}{\makecell{Database}} &  \multirow{2}{*}{\makecell{System}} &  \multicolumn{5}{c||}{Overall metrics} & \multicolumn{5}{c|}{Class-wise MF1} & \parbox{0pt}{\rule{0pt}{0.5ex+\baselineskip}} \\ [0ex]  	
				\cline{3-12}
				& & Acc. & $\kappa$ & MF1 & Sens. & Spec. & W & N1 & N2 & N3 & REM & \parbox{0pt}{\rule{0pt}{0.5ex+\baselineskip}} \\ [0ex]  	
				\cline{1-12}
				
				\multirow{7}{*}{\makecell{~\\SHHS}}  &   {\bf SleepTransformer} &  $\bm{87.7}$ & $\bm{0.828}$ & $80.1$ & $78.7$ & $\bm{96.5}$ & $92.2$ & $46.1$ & $88.3$ & $\bm{85.2}$ & $88.6$ &  \parbox{0pt}{\rule{0pt}{0ex+\baselineskip}} \\ [0ex]  	
				
				&   XSleepNet2 &  {${87.6}$} & {${0.826}$} & {${80.7}$} & {$79.7$} & {$\bm{96.5}$} &  {$92.0$} & {${49.9}$} & {${88.3}$} & {${85.0}$} & {${88.2}$}  &  \parbox{0pt}{\rule{0pt}{0ex+\baselineskip}} \\ [0ex]  	
				
				& XSleepNet1 &   ${87.5}$ & ${0.826}$ & $\bm{81.0}$ & $\bm{80.4}$ & $\bm{96.5}$ &${91.6}$ & $51.4$ & $\bm{88.5}$ & ${85.0}$ & ${88.4}$ & \parbox{0pt}{\rule{0pt}{0ex+\baselineskip}} \\ [0ex]  	

				& U-Sleep$^\dagger$ \cite{Perslev2021} &  ${-}$ & $-$ & $80.0$ & $-$ & $-$ & $93.0$ & $\bm{51.0}$ & ${87.0}$ & $76.0$ & $\bm{92.0}$ & \parbox{0pt}{\rule{0pt}{0ex+\baselineskip}} \\ [0ex]  	
				
				& Olesen \emph{et al.}$^\dagger$ \cite{Olesen2021}&   ${87.1}$ & $0.816$ & $78.8$ & $77.7$ & $96.3$ &$\bm{94.1}$ & $47.8$ & $87.9$ & $74.3$ & $89.9$ & \parbox{0pt}{\rule{0pt}{0ex+\baselineskip}} \\ [0ex]  	
				
				& SeqSleepNet \cite{Phan2019a} &   $ 86.5$ & $ 0.811$ & $ 78.5$ & $ 76.9$ & $ 96.1$ & $ 91.4$ & $ 43.3$ & $ 87.4$ & $ 82.9$ & $ 87.3$ &  \parbox{0pt}{\rule{0pt}{0ex+\baselineskip}} \\ [0ex]  	
				& FCNN+RNN &   $ 86.7$ & $ 0.813$ & $ 79.5$ & $ 78.1$ & $ 96.2$ & $ 91.1$ & $ 48.7$ & $ 88.0$ & $ 82.6$ & $ 87.1$ &  \parbox{0pt}{\rule{0pt}{0ex+\baselineskip}} \\ [0ex]  	
				
				& CNN \cite{Sors2018} &   $86.8$ & $0.810$ & $78.5$ & $-$ & $95.0$ & $-$ & $-$ & $-$ & $-$ & $-$ &  \parbox{0pt}{\rule{0pt}{0ex+\baselineskip}} \\ [0ex]  	
				&  IITNet \cite{Seo2020} & $86.7$ & $0.810$ & $79.8$ & $-$ & $-$ & $-$ & $-$ & $-$ & $-$ & $-$ &  \parbox{0pt}{\rule{0pt}{0ex+\baselineskip}} \\ [0ex]  	
				
				& AttnSleep$^\dagger$ \cite{Eldele2021} &   $84.2$ & $0.78$ & $75.3$ & $-$ & $-$ & $86.7$ & $33.2$ & $87.1$ & $87.1$ & $82.1$  &   \parbox{0pt}{\rule{0pt}{0ex+\baselineskip}} \\ [0ex]  	
				
				\cline{1-12}
				
				\multirow{9}{*}{\makecell{~\\SleepEDF-78}} &   {\bf SleepTransformer*} &  $\bm{84.9}$ & $\bm{0.789}$ & $78.8$ & $\bm{78.2}$ & $\bm{95.9}$ & $\bm{93.5}$ & ${48.5}$ & $\bm{86.5}$ & $\bm{80.9}$ & ${84.6}$  & \parbox{0pt}{\rule{0pt}{0ex+\baselineskip}} \\ [0ex]  	
				
				& {\bf SleepTransformer} &  ${81.4}$ & ${0.743}$ & ${74.3}$ & ${74.5}$ & ${95.0}$ & ${91.7}$ & ${40.4}$ & ${84.3}$ & ${77.9}$ & ${77.2}$   & \parbox{0pt}{\rule{0pt}{0ex+\baselineskip}} \\ [0ex]  	

				&   XSleepNet2 &  {${84.0}$} & {${0.778}$} & {${77.9}$} & {${77.5}$} & {${95.7}$} &  {${93.3}$} & {${49.9}$} & {$86.0$} & {$78.7$} & {$81.8$}  & \parbox{0pt}{\rule{0pt}{0ex+\baselineskip}} \\ [0ex]  	
				
				&   XSleepNet1 &   {${83.6}$} & {${0.773}$} & {${77.8}$} & {${77.7}$} & {${95.7}$} & {$92.6$} & {${50.2}$} & {$85.9$} & {$79.2$} & {$81.3$} &  \parbox{0pt}{\rule{0pt}{0ex+\baselineskip}} \\ [0ex]  	
				
				&   SeqSleepNet\cite{Phan2019a} &   $ 82.6$ & $ 0.760$ & $ 76.4$ & $ 76.3$ & $ 95.4$ & $ 92.2$ & $ 47.8$ & $ 84.9$ & $ 77.2$ & $ 79.9$ &   \parbox{0pt}{\rule{0pt}{0ex+\baselineskip}} \\ [0ex]  	
				&   FCNN+RNN &   $ 82.8$ & $ 0.761$ & $ 76.6$ & $ 75.9$ & $ 95.4$ & $ 92.5$ & $ 47.3$ & $ 85.0$ & $ 79.2$ & $ 78.9$ &  \parbox{0pt}{\rule{0pt}{0ex+\baselineskip}} \\ [0ex]  	
				& Zhu \emph{et al.} \cite{Zhu2020} &   $82.8$ & $-$ & $77.8$ & $-$ & $-$ & $90.3$ & $47.1$ & $86.0$ & $82.1$ & $83.2$  &   \parbox{0pt}{\rule{0pt}{0ex+\baselineskip}} \\ [0ex]  	
				
				&   U-Time \cite{Perslev2019} &   $-$ & $-$ & $76.0$ & $-$ & $-$ & $-$ & $-$ & $-$ & $-$ & $-$ &    \parbox{0pt}{\rule{0pt}{0ex+\baselineskip}} \\ [0ex]  	
				
				& U-Sleep$^\dagger$ \cite{Perslev2021} &  ${-}$ & $-$ & $\bm{79.0}$ & $-$ & $-$ & $93.0$ & $\bm{57.0}$ & ${86.0}$ & $71.0$ & $\bm{88.0}$ & \parbox{0pt}{\rule{0pt}{0ex+\baselineskip}} \\ [0ex]  	
				
				&   CNN-LSTM \cite{Perslev2019} &   $-$ & $-$ & $73.0$ & $-$ & $-$ &  $-$ & $-$ & $-$ & $-$ & $-$ &    \parbox{0pt}{\rule{0pt}{0ex+\baselineskip}} \\ [0ex]  	
				
				& AttnSleep \cite{Eldele2021} &   $81.3$ & $0.74$ & $75.1$ & $-$ & $-$ & $92.0$ & $42.0$ & $85.0$ & $82.1$ & $74.1$  &   \parbox{0pt}{\rule{0pt}{0ex+\baselineskip}} \\ [0ex]  	
				
				& SleepEEGNet \cite{MousaviI2019} &   $80.0$ & $0.730$ & $73.6$ & $-$ & $-$ & $-$ & $-$ & $-$ & $-$ & $-$ &   \parbox{0pt}{\rule{0pt}{0ex+\baselineskip}} \\ [0ex]  	
				\cline{1-12}

				
				
				
				

			\end{tabular}
		\end{center}
		\label{tab:performance}
		\vspace{-0.3cm}
	\end{table*}

	\section{Experiments}
	\label{sec:experiments}
	\subsection{Extraction of time-frequency image}	
	\label{ssec:featureextraction}
	
	As described, SleepTransformer ingests time-frequency images as input. To extract a time-frequency image, a 30-second EEG epoch sampled at 100 Hz was decomposed into two-second frames with 50\% overlap, multiplied with a Hamming window, and transformed to the frequency domain using a 256-point short-time Fourier Transform (STFT). This procedure resulted in an image $\mathbf{S} \in \mathbb{R}^{T\times F}$ with $T\!=\!29$ time frames and $F\!=\!128$ frequency bins. Of note, we excluded the $0$-th frequency bin to keep $F\!=\!128$ which is a multiple of the number of attention heads in the epoch transformer in Section \ref{ssec:epochtransformer}. The amplitude spectrum was then log-transformed. The time-frequency images extracted from a database were normalized to zero mean and unit variance along the frequency dimension given the normalization parameters computed using the training data.

	\subsection{Parameters}	
	
	We experimented with different values $\{11, 21, 31, 41, 51\}$ for the input sequence length $L$ and 21 was found to be best. This result confirms the finding reported in other works like \cite{Phan2019a,Supratak2017}. Thus, we fixed $L\!=\!21$ for further experiments here. The network was designed to have $N_E\!=\!4$ \emph{EpochTransformer}s for intra-epoch processing and $N_S\!=\!4$ \emph{SequenceTransformer}s for inter-epoch processing. 
	In a transformer, the number of attention heads was fixed to $H\!=\!8$ and the number of hidden units of a feed forward layer was fixed to $d_{FF}=1024$. The FC layers of the network was also of size 1024. A common dropout rate of 0.1 was applied to the transformer, including the self-attention layers and the feed forward layers, as well as the FC layers.
	
	The experiments were conducted on the two databases SHHS and SleepEDF-78 individually. We carried out 10-fold cross-validation on the SleepEDF-78 database as in prior works \cite{Phan2021c,Phan2020a,MousaviI2019}. In each iteration, seven subjects were left out from the training subjects as the validation set. For the large-scale database, SHHS, we randomly split the subject into 70\% for training and 30\% for testing as in \cite{Phan2021c,Sors2018}. 100 subjects were left out from the training set as the validation set. Of note, following \cite{Sors2018}, those recordings without all five sleep stages were excluded. The network was trained using Adam optimizer \cite{Kingma2015} with a learning rate of $10^{-4}$, $\beta_1\!=\!0.9$,  $\beta_2\!=\!0.999$, and $\epsilon\!=\!10^{-7}$. A minibatch size of 32 was used for training. The model was validated on the validation set every 100 training steps. Early stopping was applied and the training was stopped after 200 validation steps without improvement on the validation data. Particularly, for SHHS, the model was trained for at least 5000 validation steps before early stopping was activated.
	
	\vspace{-0.2cm}
	\subsection{Experimental results}	
	\subsubsection{Sleep staging performance}
	
	Table \ref{tab:performance} shows the performance on the experimental databases obtained by SleepTransformer in comparison to those reported in previous works. Accuracy, Cohen's kappa ($\kappa$) \cite{McHugh2012}, macro F1-score (MF1) \cite{Yang1999}, average sensitivity, and average specificity are used as overall performance metrics while class-specific performance is assessed using class-wise MF1. 
	
	On the large-scale database, SHHS, SleepTransformer achieves an overall accuracy of $87.7\%$ and $\kappa$ of $0.828$. On the one hand, compared to the seminal sequence-to-sequence counterpart, SeqSleepNet \cite{Phan2019a}, SleepTransformer results in an improvement of $1.2\%$ absolute in terms of accuracy and $0.017$ in terms of $\kappa$. This result suggests that the transformer backbone is more advantageous than the recurrent backbone used in SeqSleepNet. On the other hand, SleepTransformer's performance is on par with those of the existing state-of-the-art XSleepNets \cite{Phan2021c} (i.e. XSleepNet1 and XSleepNet2 in Table \ref{tab:performance}). 
	Although the differences in their accuracy and $\kappa$ are small, it is considerable given the fact that SleepTransformer only uses the time-frequency input (i.e. single view), has a smaller model footprint, and is computationally cheaper (see Table \ref{tab:modelsize_trainingtime}). The class-wise MF1s further unravel their opposite patterns. SleepTransformer seems to favor the major sleep stages, i.e. Wake and REM, over the under-present stage N1. On the contrary, the class-wise MF1s are less skewed in case of XSleepNets, resulting in a better overall MF1 than SleepTransformer.
	
	On the smaller database (SleepEDF-78) SleepTransformer's performance seems to be inferior to other competitors. However, this result does not correctly reflect its modelling capacity. In fact, we observed that SleepTransformer overfitted this database quite easily and thus requires a larger amount of data for training. Motivated by the sleep transfer learning approach in \cite{Phan2020b}, we utilized the model trained on SHHS to initialize the network instead of random initialization. This simple trick significantly boosted the performance, improving the overall absolute accuracy, $\kappa$, and MF1 by $3.5\%$, $0.046$, and $4.5\%$. With the achieved overall accuracy of $84.9\%$, $\kappa$ of $0.789$, and MF1 of $78.8$, SleepTransformer outperforms all the previous works evaluated on SleepEDF-78, except for U-Sleep \cite{Perslev2021} which utilized an ensemble of models separately trained on EEG and EOG. Regarding the class-wise performance, similar conclusions as for SHHS can be drawn.
	
	\subsubsection{Confidence estimation}
	
	\begin{figure} [!t]
		\centering
		\includegraphics[width=0.95\linewidth]{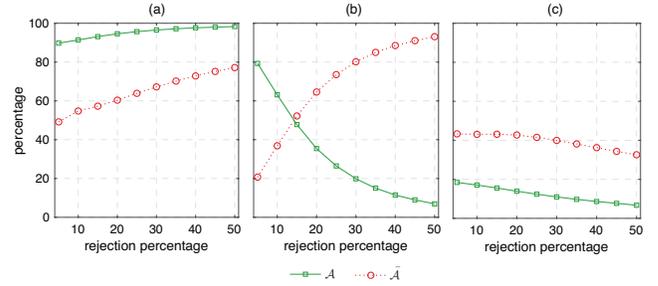}
		\vspace{-0.1cm}
		\caption{SleepEDF-78: {\bf (a)} Accuracies of $\mathcal{A}$ and $\mathcal{\bar{A}}$; {\bf (b)} percentages of misclassified epochs in $\mathcal{A}$ and $\mathcal{\bar{A}}$ out of all misclassified epochs; {\bf (c)} percentages of transitioning epochs out of $\mathcal{A}$ and $\mathcal{\bar{A}}$.}
		\label{fig:accuracy_vs_rejection_percentage}
		\vspace{-0.2cm}
	\end{figure}
	
	\begin{figure*} [!t]
		\centering
		\includegraphics[width=0.9\linewidth]{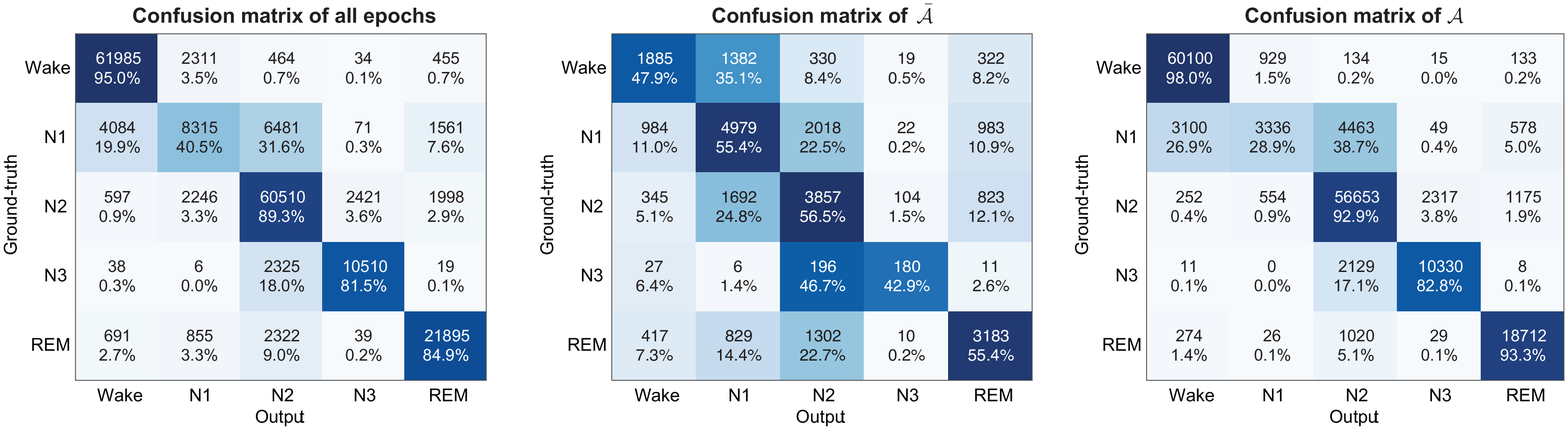}
		\vspace{-0.1cm}
		\caption{Confusion matrices of SleepEDF-78. A confidence threshold of 0.5 was used to separate the epochs into $\mathcal{A}$ and $\mathcal{\bar{A}}$.}
		\label{fig:confusion_matrices}
		\vspace{-0.2cm}
	\end{figure*}
	
	\begin{figure*} [!t]
		\centering
		\includegraphics[width=0.95\linewidth]{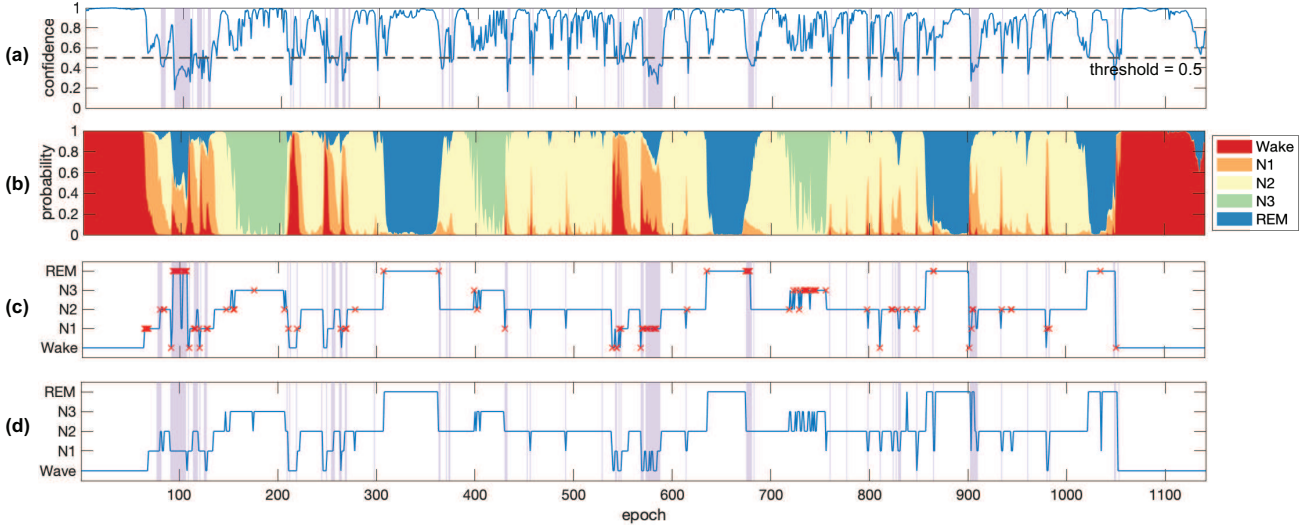}
		\vspace{-0.2cm}
		\caption{Visualization the estimated confidence for Subject 2 of SleepEDF-78. {\bf (a)} The quantified confidence; {\bf (b)} the probability output; {\bf (c)} the output hypnogram where \textcolor{red}{$\times$} indicates the misclassified epochs; and {\bf (d)} the ground-truth hypnogram. The shaded regions correspond to the epochs with their confidence below the threshold.}
		\label{fig:confidence}
		\vspace{-0.2cm}
	\end{figure*}
	
	Let $\mathcal{\bar{A}}$ denote the set of low-confidence epochs (that need further manual verification and/or correction) whose confidences are below a predefined threshold.  Alternatively, $\mathcal{\bar{A}}$ can also be selected as the set of epochs with lowest confidences, e.g. the set containing 20\% of epochs with lowest confidences. In addition, let $\mathcal{A}$ denote the set of the remaining epochs, which is the complement of $\mathcal{\bar{A}}$.
	
	Reasonably, the confidence metric is only useful and meaningful if it could help filter out misclassified epochs for further manual verification/correction. As shown in Figure \ref{fig:accuracy_vs_rejection_percentage} (a), the classification accuracy of $\mathcal{\bar{A}}$ remains much lower than $\mathcal{A}$ regardless of the size of $\mathcal{\bar{A}}$. For example, when $\mathcal{\bar{A}}$ constitutes 20\% of all the epochs, it has an accuracy around 60\%, meaning 40\% of its epochs are misclassified ones. And these misclassified epochs constitute about 65\% of all misclassified epochs as shown in Figure \ref{fig:accuracy_vs_rejection_percentage} (b). The percentage of misclassified epochs in $\mathcal{\bar{A}}$ increases sharply when it grows larger. When $\mathcal{\bar{A}}$ constitutes of 50\% of all the epochs, we can segregate more than 90\% of all misclassified epochs. All of this implies that the misclassified epochs are often associated with low confidences. For sleep particularly, the transitioning epochs (whose sleep stages are different from those of its preceding and/or succeeding neighbors \cite{Phan2019b}) are usually difficult ones as human scorers also tend to disagree and machine scoring systems often make mistakes. Interestingly, the percentage of transitioning epochs in $\mathcal{\bar{A}}$ is always considerably larger than in $\mathcal{A}$, as shown in Figure \ref{fig:accuracy_vs_rejection_percentage} (c). Moreover, the confusion matrices in Figure \ref{fig:confusion_matrices} further reveal that the majority of the epochs in $\mathcal{\bar{A}}$ are N1, leaving just a small portion of N1 epochs in $\mathcal{A}$. This is not a surprise since N1 is, in general, hard to be correctly recognized due to its under-presence and strong resemblance to Wake and N2. 
	
\begin{figure} [!t]
		\centering
		\includegraphics[width=1\linewidth]{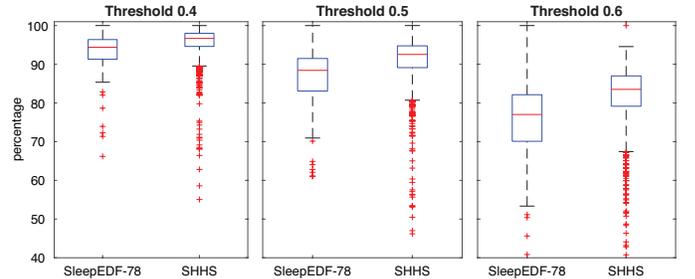}
		\caption{Percentages of epochs per night with confidence above a confidence threshold of 0.4, 0.5, and 0.6.}
		\label{fig:confidence_threshold}
	\end{figure}

	We further showcase the above findings in Figure \ref{fig:confidence} where we portray the quantified confidence alongside the ground-truth hypnogram, the output hypnogram, and the multi-class probability output for a subject in SleepEDF-78 (i.e. Subject 2). In the figure, the confidence is thresholded by 0.5. It can be seen that, often, the low-confidence and transitioning epochs are misclassified. With the threshold value 0.5, roughly 90\% (in case of SleepEDF-78) or more (in case of SHHS) of epochs per night have their confidences above the threshold, as further shown in Figure \ref{fig:confidence_threshold}, and would not need to undergo manual verification.
	
	\subsubsection{Attention score visualization for interpretation}
	
	\begin{figure*} [!t]
		\centering
		\includegraphics[width=0.95\linewidth]{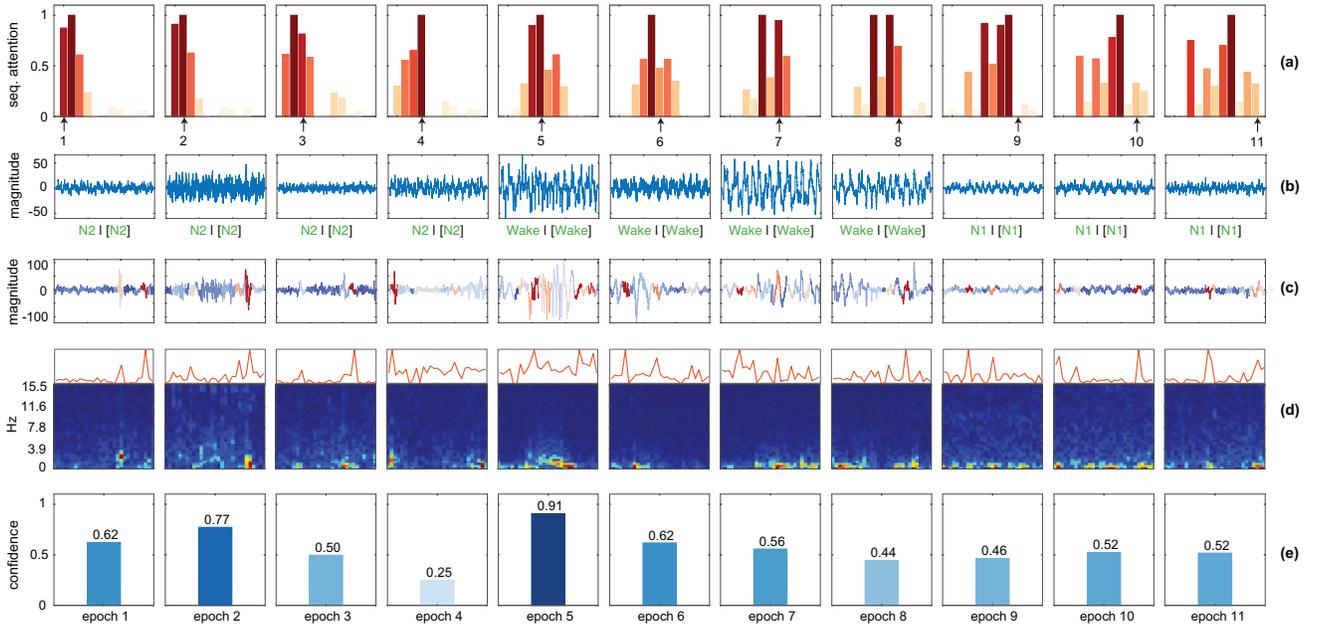}
		\caption{Self-attention visualization for interpretation of an input sequence with $L=11$ stemming from SleepEDF-78 (Subject 2). The sequence involves two cross-stage transitions: (i) N2$\rightarrow$Wake around epochs 4 and 5, and (ii) Wake$\rightarrow$N1 around epochs 8 and 9. {\bf (a)} The distribution of sequence-level attention scores where an arrow indicates the position of the current epoch in the distribution of the attention score. {\bf (b)} The EEG features learned for epoch representation (constructed via ISTFT as described in Section \ref{ssec:explainability}). For pair of labels, the output label is on the left and the ground-truth label is inside the brackets on the right. {\bf (c)} The epoch-level attention scores represented by heat map on the raw EEG signals. {\bf (d)} The spectrogram inputs and their epoch attention scores (i.e. the red curves) distributed over spectral columns. {\bf (e)} The estimated confidences.}
		\label{fig:interpretation1}
	\end{figure*}
	\begin{figure*} [!t]
		\centering
		\includegraphics[width=0.95\linewidth]{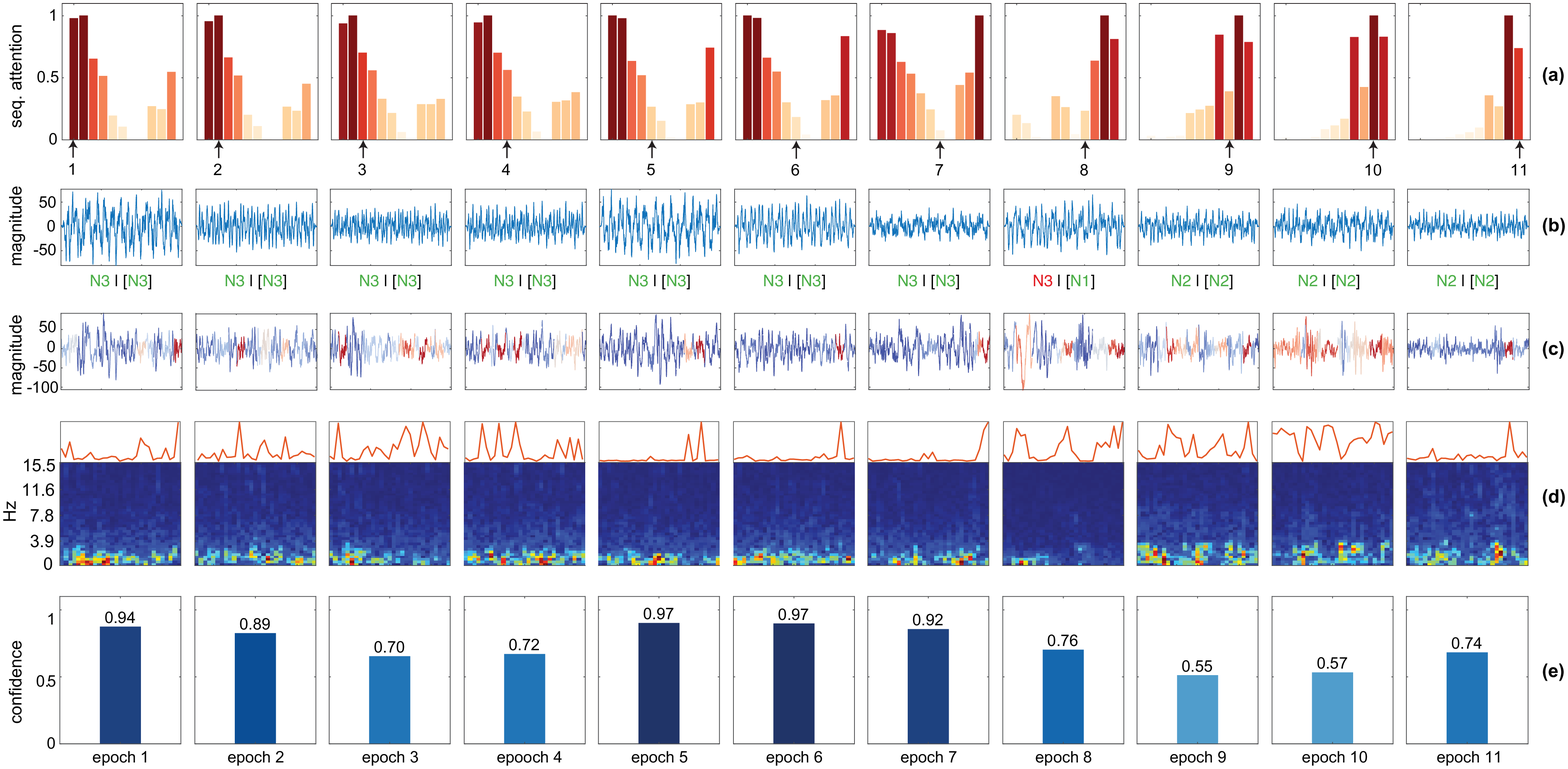}
		\caption{Self-attention visualization for interpretation of another input sequence with $L=11$ stemming from SleepEDF-78 (Subject 2). The sequence involves two cross-stage transitions: (i) N3$\rightarrow$N1 around epochs 7 and 8, and (ii) N1$\rightarrow$N2 around epochs 8 and 9. {\bf (a)} The distribution of sequence-level attention scores where an arrow indicates the position of the current epoch in the distribution of the attention score. {\bf (b)} The EEG features learned for epoch representation (constructed via ISTFT as described in Section \ref{ssec:explainability}). For pair of labels, the output label is on the left and the ground-truth label is inside the brackets on the right. {\bf (c)} The epoch-level attention scores represented by heat map on the raw EEG signals. {\bf (d)} The spectrogram inputs and their epoch attention scores (i.e. the red curves) distributed over spectral columns. {\bf (e)} The estimated confidences.}
		\label{fig:interpretation2}
	\end{figure*}
	
	\setlength\tabcolsep{2.25pt}
	\begin{table*}[!t]
		\caption{Variation of SleepTransformer's performance on SHHS with different values of $N_E$, the number of transformer encoder block in the epoch encoder, and $N_S$, the number of transformer encoder block in the sequence encoder.}
		\vspace{-0.2cm}
		\begin{center}
			\begin{tabular}{|>{\centering\arraybackslash}m{0.35in}|>{\arraybackslash}m{0.35in}|>{\centering\arraybackslash}m{0.4in}|>{\centering\arraybackslash}m{0.45in}|>{\centering\arraybackslash}m{0.4in}|>{\centering\arraybackslash}m{0.4in}|>{\centering\arraybackslash}m{0.4in}||>{\centering\arraybackslash}m{0.4in}|>{\centering\arraybackslash}m{0.45in}|>{\centering\arraybackslash}m{0.4in}|>{\centering\arraybackslash}m{0.4in}|>{\centering\arraybackslash}m{0.4in}|>{\centering\arraybackslash}m{0in} @{}m{0pt}@{}}
				\cline{1-12}
				\multirow{2}{*}{$N_E$} &  \multirow{2}{*}{$N_S$} &  \multicolumn{5}{c||}{Overall metrics} & \multicolumn{5}{c|}{Class-wise MF1} & \parbox{0pt}{\rule{0pt}{1ex+\baselineskip}} \\ [0ex]  	
				\cline{3-12}
				& & Acc. & $\kappa$ & MF1 & Sens. & Spec. & W & N1 & N2 & N3 & REM & \parbox{0pt}{\rule{0pt}{0.5ex+\baselineskip}} \\ [0ex]  	
				\cline{1-12}
				1 & 4  & $87.5$ & $0.824$ & $79.3$ & $77.8$ & $96.4$ & $91.9$ & $42.8$ & $88.2$ & $85.1$ & $88.2$ & \parbox{0pt}{\rule{0pt}{0ex+\baselineskip}} \\ [0ex]  	
				2 & 4  &  $\bm{87.7}$ & $0.827$ & $79.5$ & $77.9$ & $\bm{96.5}$ & $92.3$ & $42.9$ & $\bm{88.3}$ & $85.2$ & $88.6$ &\parbox{0pt}{\rule{0pt}{0ex+\baselineskip}} \\ [0ex]  	
				3 & 4  & $\bm{87.7}$ & $\bm{0.828}$ & $79.8$ & $78.4$ & $\bm{96.5}$ & $\bm{92.4}$ & $44.2$ & $\bm{88.3}$ & $85.1$ & $\bm{88.9}$ & \parbox{0pt}{\rule{0pt}{0ex+\baselineskip}} \\ [0ex]  	
				\cline{1-12}
				4 & 4  &  $\bm{87.7}$ & $\bm{0.828}$ & $\bm{80.1}$ & $\bm{78.7}$ & $\bm{96.5}$ & $92.2$ & $\bm{46.1}$ & $\bm{88.3}$ & $85.2$ & $88.6$ &\parbox{0pt}{\rule{0pt}{0ex+\baselineskip}} \\ [0ex]  	
				\cline{1-12}
				4 & 3  &  $87.4$ & $0.823$ & $79.1$ & $77.7$ & $96.4$ & $92.1$ & $42.3$ & $87.9$ & $85.1$ & $88.1$ & \parbox{0pt}{\rule{0pt}{0ex+\baselineskip}} \\ [0ex]  	
				4 & 2  &  $87.6$ & $0.826$ & $79.6$ & $78.5$ & $\bm{96.5}$ & $\bm{92.4}$ & $44.0$ & $88.2$ & $85.2$ & $88.3$ & \parbox{0pt}{\rule{0pt}{0ex+\baselineskip}} \\ [0ex]  	
				4 & 1  &  $87.5$ & $0.825$ & $79.1$ & $77.9$ & $\bm{96.5}$ & $92.1$ & $41.9$ & $\bm{88.3}$ & $\bm{85.5}$ & $87.7$ & \parbox{0pt}{\rule{0pt}{0ex+\baselineskip}} \\ [0ex]  	
				\cline{1-12}
			\end{tabular}
		\end{center}
		\label{tab:number_of_transformer}
		\vspace{-0.2cm}
	\end{table*}

	In light of the proposed approach for explainability in Section \ref{ssec:explainability}, in Figures \ref{fig:interpretation1} and \ref{fig:interpretation2}, we attempted to visualize the attention scores for two different input sequences at both the epoch and sequence level. For simplicity and clarity, we used the model with sequence length $L=11$ and the attention scores of the last \emph{EpochTransformer} and \emph{SequenceTransformer} for the EEG heat map and the sequence-level attention, respectively. We also included the predicted labels, the ground-truth labels, and the estimated confidences of the epochs in the sequences to aid interpretability. We additionally show enlarged versions of the raw EEG signals in Appendix for further details. 
	
	
	
	At the epoch-level, the heat map on the EEG signals in the figures suggests that the model indeed attended more to sleep related features. For instance, the K-complexes present in epochs 2 and 4 in Figure \ref{fig:interpretation1}(b). 
	This type of micro-event is notable in the sleep stage N2. The attention is more scattering for N3 stage in epochs 1, 2, and 3 in Figure \ref{fig:interpretation1}(b) 
	given the omnipresence of Delta waves. Furthermore, the constructed EEGs resemble Alpha waves in N1 stage (epochs 9, 10, and 11 in Figure \ref{fig:interpretation1}(c)),
	high-amplitude neural activities in Wake stage (epochs 5, 7, and 8 in Figure \ref{fig:interpretation1}(c)),
	or Delta waves in N3 stage (epochs 1-6 in Figure \ref{fig:interpretation2}(c)).
	These constructed EEGs exhibit distinguishable frequency distributions as shown in Figures \ref{fig:wake_tf} and \ref{fig:N3_tf}.

	At the sequence level, the attention scores act as the weights used to collectively combine features in different epochs in the sequence to classify a target epoch, featuring the benefit of sequence-to-sequence sleep scoring with self-attention. In the first example in Figure \ref{fig:interpretation1}, apparently, those epochs on the left of the sequence containing useful features for recognizing the stage N2 are associated with strong weights whereas other epochs containing less relevant features are associated with smaller weights. Similarly, for Wake epochs, the attention scores concentrate more around the epochs in the middle. On the contrary, the attention scores of the N1 epochs on the right disperse due to the fact that the stage N1 shares similar features to both Wake and N2. In other words, even though the EEG signal of a target epoch does not show much useful features for recognizing the sleep stage N1, the model is still able to recognize it by leveraging the relevant features appearing in the context via the sequence-level attention. 
	
	In the second example in Figure \ref{fig:interpretation2},
	we note that, for both N2 and N3, considerably greater attention weights are placed on the epochs far away from the transitioning boundary which most likely convey more reliable features. 
	Thus, the network is taking advantage of long-term structure in sleep to recognize those epochs close to the transitioning boundary. This particular example, on the other hand, demonstrates an interesting case at the 8-th epoch where the network misclassified, predicting N3 against the ground-truth N1. We argue that the ground-truth N1 in this case highlights the well-known subjectivity of human scoring since the transient transition N3$\rightarrow$N1$\rightarrow$N2 seems to be counter-intuitive. This epoch seems to contain mixed information of both N3 (delta activity as shown in the time-frequency image for the 8-th epoch in Figure \ref{fig:N3_tf}) and N2 (the big K-complex in the raw EEG of the 8-th epoch in Figure \ref{fig:interpretation2} (c)). Thus, it is more likely to be a N3$\rightarrow$N2 transitioning epoch.
	
	
	
	We argue that this visualization resembles the way human scoring is done, and therefore, would facilitate manual verification and correction of low-confidence epochs and provide a gateway for practitioners to interact with the model.
	
	\vspace{-0.2cm}
	\subsection{Discussion}	
	
	For models using the transformer backbone like SleepTransformer, choosing an appropriate number of transformers can be crucial. To investigate the influence of the number of \emph{EpochTransformer} $N_E$ and the number of \emph{SequenceTransformer} $N_S$, we repeated the experiments with $N_E$ fixed to $4$ and $N_S$ varied in $\{1, 2, 3, 4\}$. After that, we repeated the experiments with $N_S$ fixed to 4 and $N_E$ varied in $\{1, 2, 3, 4\}$. Using SHHS for this investigation, the overall performance obtained with different values of  $N_E$ and $N_S$ are shown in Table \ref{tab:number_of_transformer}. These results suggest the modest impact of both $N_E$ and $N_S$ on the overall accuracy. However, the class-wise MF1s suggest a large number of transformers is important to improve the performance (mostly the sensitivity, i.e. the true positive rate) on the under-present stage N1 which, in turn, improves the average MF1.

	Concerning the model size and computational cost, even with $N_E\!=\!4$ and $N_S\!=\!4$, SleepTransformer has moderate model size and modest computational overhead as contrasted with some existing models (whose relevant information was previously reported) in Table \ref{tab:modelsize_trainingtime}.
	In particular, compared to our recent developed model, XSleepNet \cite{Phan2021c}, SleepTransformer's model size is just two third of it of XSleepNet \cite{Phan2021c} while it is 2.7 times faster to train. It is even faster than SeqSleepNet, the compact model proposed in our previous work \cite{Phan2019a}, most likely because SleepTransformer is recurrent-free. Of note, we measured the training time of the models in the table using a common DGX-2 machine with NVIDIA Tesla V100 graphic card and Intel Xeon Platinum 8186 CPU, 2.7 GHz. 
	
	Future work can further address the following limitations of this work. First, the entropy-based uncertainty quantification proposed here is not only applicable for SleepTransformer but also any sleep-scoring model with probability output, such as SeqSleepNet\cite{Phan2019a} or XSleepNet \cite{Phan2021c}. Furthermore, alternative to entropy, the likelihood of the most likely class (i.e. the maximum probability in an output probability distribution) could also be used to obtain a measure of uncertainty \cite{Mikkelsen2019b}. Different from entropy, which depends on the entire distribution over classes to measure an overall uncertainty in the predictions, this measure of uncertainty is not affected by probabilities of other classes and may yield a more precise estimation. Lastly, we are only dealing with \emph{knowledge uncertainty} (i.e. uncertainty in the model's predictions), leaving \emph{data uncertainty} \cite{Mikkelsen2020} (i.e. uncertainty arises due to the complexity, multi-modality and noise in the data) open for future works. Ideally, a method that could take into account both data uncertainty and model uncertainty, could be agnostic to the network architectures, and could be applied to already trained models would be much more useful. Second, our visualization attempt in Figure \ref{ssec:explainability} is not necessarily the best and the only way to interpret the model's decisions. Further creativity and interaction with experts will be needed to leverage the information encoded in the attention scores before it can be embedded in daily sleep practise. Third, we employed the original transformer proposed in the seminal work of \cite{Vaswani2017} as the backbone of SleepTransformer, more advanced variants of transformer can be further explored.
	
		\begin{table}[!t]
		\caption{Model size and training time per 1000 training steps.}
		\vspace{-0.3cm}
		\begin{center}
			\begin{tabular}{|>{\arraybackslash}m{1in}|>{\centering\arraybackslash}m{0.75in}|>{\centering\arraybackslash}m{1in}|>{\centering\arraybackslash}m{0in} @{}m{0pt}@{}}
				\cline{1-3}
				\makecell{Model} & \makecell{\#parameters} & \makecell{training time (s)\\per 1000 steps} & \parbox{0pt}{\rule{0pt}{0.5ex+\baselineskip}} \\ [0ex]  	
				\cline{1-3}
				SleepTransformer & $3.70 \times 10^6$  & 308 & \parbox{0pt}{\rule{0pt}{0.5ex+\baselineskip}} \\ [0ex]  	
				
				SeqSleepNet \cite{Phan2019a} & $1.64 \times 10^5$  & 379 & \parbox{0pt}{\rule{0pt}{0ex+\baselineskip}} \\ [0ex]  	
				
				U-Time \cite{Perslev2019} & $1.10 \times 10^6$ & $-$ & \parbox{0pt}{\rule{0pt}{0ex+\baselineskip}} \\ [0ex]  	
				
				U-Sleep \cite{Perslev2021} & $3.10 \times 10^6$  & $-$ & \parbox{0pt}{\rule{0pt}{0ex+\baselineskip}} \\ [0ex]  	
				
				XSleepNet \cite{Phan2021c} & $5.74\times 10^6$  &  828 &\parbox{0pt}{\rule{0pt}{0ex+\baselineskip}} \\ [0ex]  	
				
				DeepSleepNet \cite{Supratak2017} & $2.30 \times 10^7$ & $-$ & \parbox{0pt}{\rule{0pt}{0ex+\baselineskip}} \\ [0ex]  	

				\cline{1-3}
			\end{tabular}
		\end{center}
		\label{tab:modelsize_trainingtime}
		\vspace{-0.2cm}
	\end{table}

	\section{Conclusions}
	\label{sec:conclusions}
	
	We proposed SleepTransformer, a sequence-to-sequence sleep staging model relying solely on the transformer network. We showed that SleepTransformer performed comparably to state-of-the-art models on both SHHS, a large-scale database, and SleepEDF-78, a relative small database. We leveraged the attention scores of the transformer's self-attention module for interpretability. At the epoch level, the attention scores was applied to the EEG input as a heat map to highlight sleep-relevant features the model attended to. At the sequence level, the attention scores were interpreted as the contribution of different neighboring epochs to the recognition of a target epoch in the input sequence. We also used entropy of the multi-class probability distribution output to quantify uncertainty of the model's decisions as a concrete number which was shown to align well with the model's mistakes and successes. 
	

	\section*{Acknowledgment}
	This research received funding from the Flemish Government (AI Research Program). Maarten De Vos is affiliated to Leuven.AI - KU Leuven institute for AI, B-3000, Leuven, Belgium. H. Phan is supported by a Turing Fellowship under the EPSRC grant EP/N510129/1. The study was approved by Clinical Trials and Research Governance, Churchill Hospital - Oxford University Hospitals, Oxford, UK. Data were provided by the Center for Sleep and Wake Disorders at MCH Westeinde Hospital, Den Haag, the Netherlands; and the Division of Sleep and Circadian Disorders, Brigham and Women’s Hospital, MA, USA.
	
	
	\bibliographystyle{IEEEbib}
	\bibliography{bibliography}
	
	\appendices
	
	\setcounter{figure}{0}
	
	\section{Time-frequency representation corresponding to the constructed EEGs}	

	\numberwithin{equation}{section}
	\numberwithin{table}{section}
	\numberwithin{figure}{section}

\begin{figure*} [!t]
	\vspace{-0.5cm}
	\centering
	\includegraphics[width=0.9\linewidth]{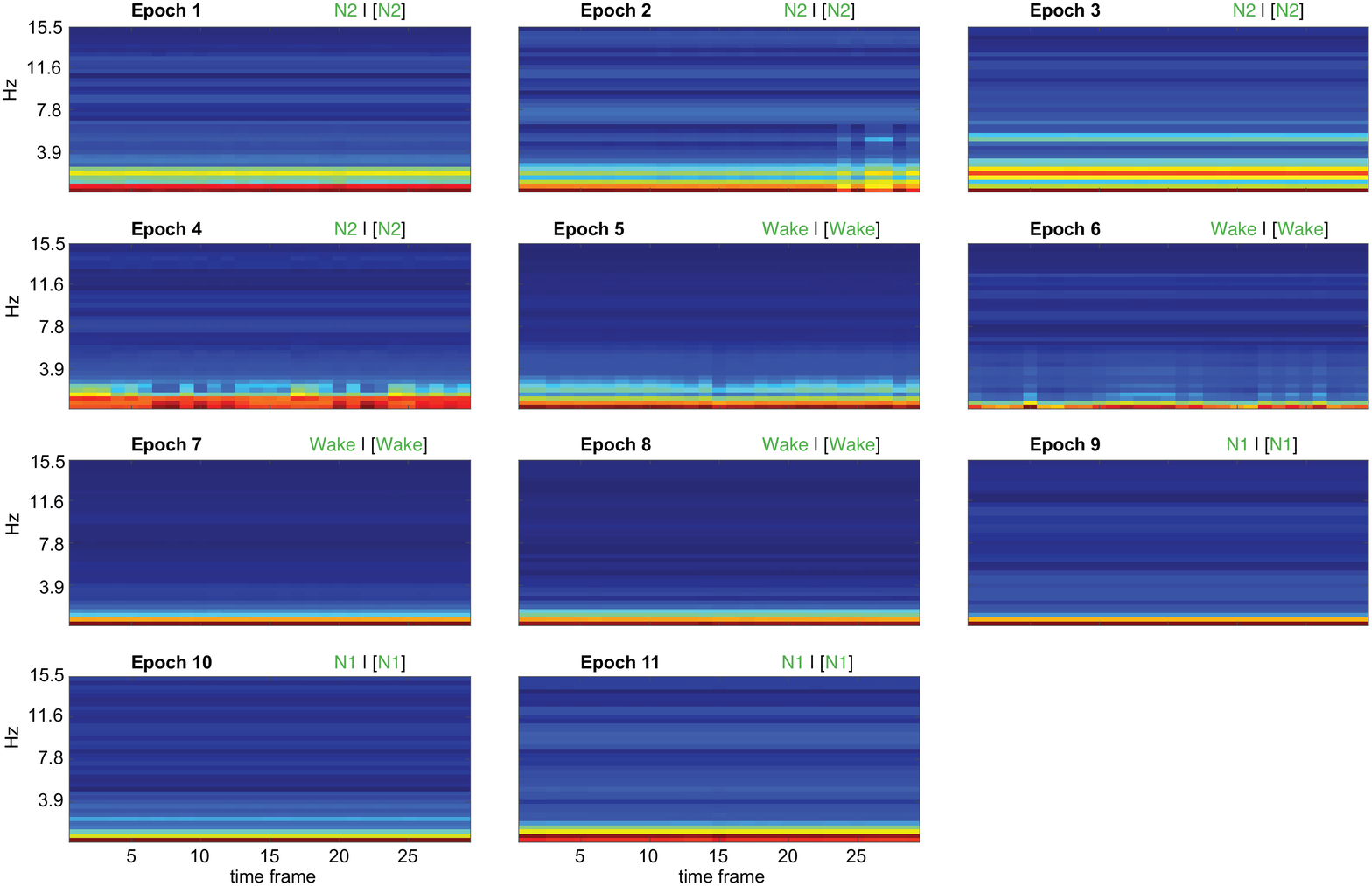}
	\caption{Time-frequency representation corresponding to the constructed EEG in Figure \ref{fig:interpretation1} (b). Of note, we only show frequency bins 1-40 corresponding to frequency range $(0, 15.5]$ Hz.}
	\label{fig:wake_tf}
	\vspace{-0.5cm}
\end{figure*}

\begin{figure*} [!t]
	\vspace{-0.5cm}
	\centering
	\includegraphics[width=0.9\linewidth]{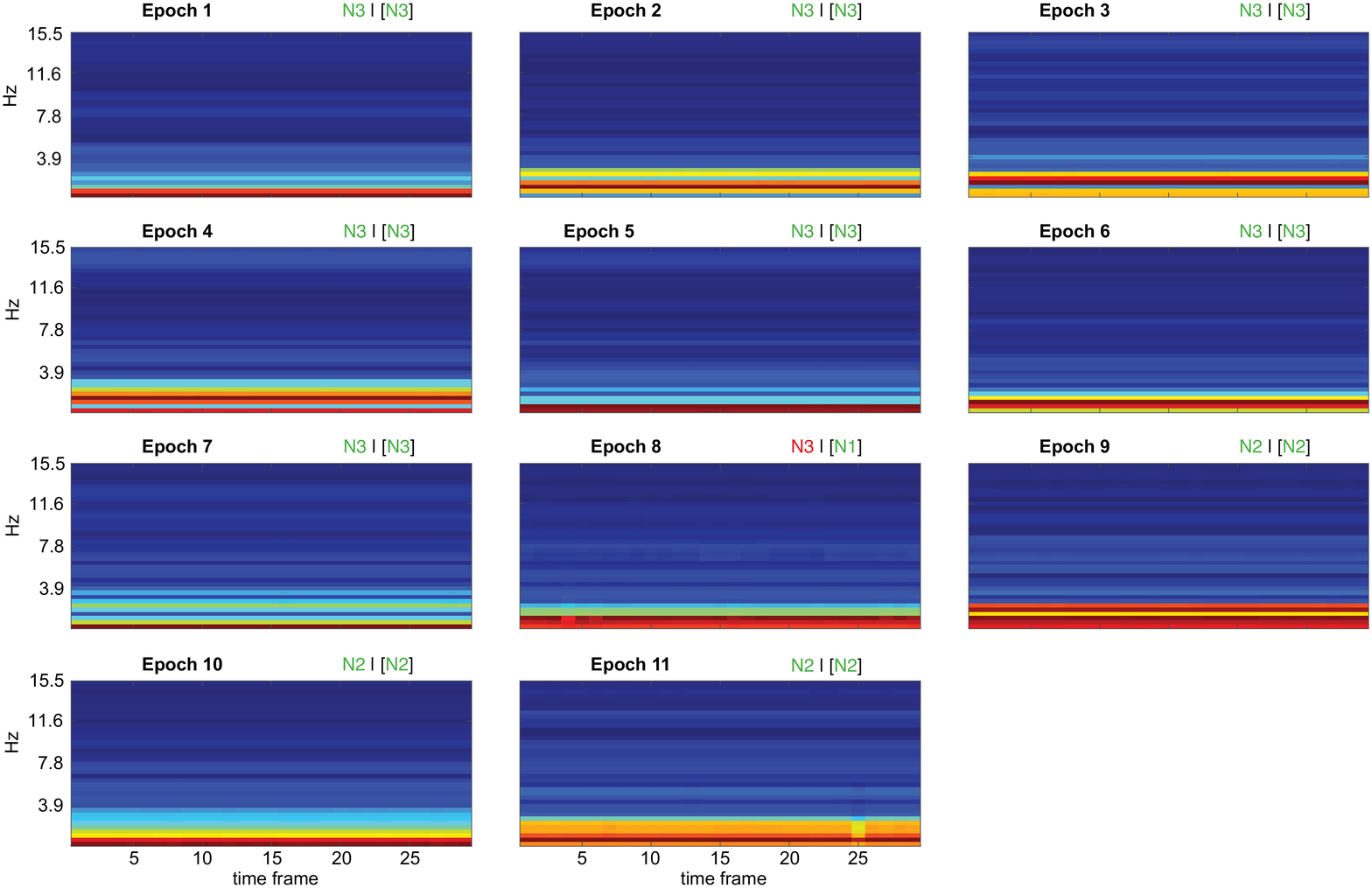}
	\caption{Time-frequency representation corresponding to the constructed EEG in Figure \ref{fig:interpretation2} (b). Of note, we only show frequency bins 1-40 corresponding to frequency range $(0, 15.5]$ Hz.}
	\label{fig:N3_tf}
\end{figure*}

\end{document}